\documentclass[lettersize,journal]{IEEEtran}
\usepackage{amsmath,amsfonts}
\usepackage{algorithmic}
\usepackage{algorithm}
\usepackage{array}
\usepackage[caption=false,font=normalsize,labelfont=sf,textfont=sf]{subfig}
\usepackage{textcomp}
\usepackage{stfloats}
\usepackage{url}
\usepackage{verbatim}
\usepackage{graphicx}
\usepackage{cite}
\begin{document}

\title{Enhanced Position Estimation in Tactile Internet-Enabled Remote Robotic Surgery Using MOESP-Based Kalman Filter}

\author{Muhammad Hanif Lashari, Wafa Batayneh, Ashfaq Khokhar,~\IEEEmembership{Fellow,~IEEE,} and Shakil Ahmed,~\IEEEmembership{Member,~IEEE}%
    \thanks{Muhammad Hanif Lashari is with the Department of Electrical and Computer Engineering, Iowa State University, Ames, IA 50011 (email: mhanif@iastate.edu).}
    \thanks{Manuscript received October 15, 2024; revised Month XX, 202X.}
}

\markboth{IEEE Transactions on Robotics,~Vol.~X, No.~X, Month~Year}%
{Lashari \MakeLowercase{\textit{et al.}}: Enhanced Position Estimation in Tactile Internet-Enabled Remote Robotic Surgery}

% IEEEpubid
%\IEEEpubid{0000--0000/00\$00.00~\copyright~2024 IEEE}

\maketitle

% Ensure the footer is properly aligned by adding some vertical space
\IEEEpubidadjcol % Moves the IEEEpubid into the second column
\vspace{10pt}    % Adds space between the footer and the tex

\begin{abstract}
Accurately estimating the position of a patient’s side robotic arm in real-time during remote surgery is a significant challenge, especially within Tactile Internet (TI) environments. This paper presents a new, efficient method for position estimation using a Kalman Filter (KF) combined with the Multivariable Output-Error State Space (MOESP) method for system identification. Unlike traditional approaches that assume prior knowledge of the system’s dynamics, this study utilizes the JIGSAW dataset—a comprehensive collection of robotic surgical data—alongside input from the Master Tool Manipulator (MTM) to derive the state-space model directly. The MOESP method allows us to accurately model the Patient Side Manipulator (PSM) dynamics without prior system models, enhancing the KF's performance under simulated network conditions, including delays, jitter, and packet loss. These conditions mimic the real-world challenges faced in Tactile Internet applications. Our findings demonstrate the KF’s enhanced resilience and accuracy in state estimation, achieving over 95\% estimation accuracy despite the presence of uncertainties induced by the network.
\end{abstract}

\begin{IEEEkeywords}
Tactile Internet, Remote Robotic Surgery, Kalman Filter, State Estimation, MOESP, JIGSAWS Dataset, PSM, MTM
\end{IEEEkeywords}


\maketitle

\section{INTRODUCTION}
\IEEEPARstart{T}{he} Tactile Internet (TI) is a cutting-edge concept that is part of the next generation of mobile communication systems, known as 6G. Super-fast and reliable networks will enable the delivery of skills and touch-based communication, leading to major societal changes. Unlike the regular internet, TI promises to offer seamless global connectivity, thanks to its advanced 6G technology. There will be different ways to interact with digital technology in the future \cite{1}.

TI, driven by groundbreaking technological advancements, focuses on real-time touch transmission using state-of-the-art haptic equipment and robotics. This innovation heralds a shift from mere content delivery to a dynamic system of skill set exchange over the internet. It promises an ultra-responsive and ultra-reliable network connectivity, crucial for applications where real-time control and feedback are imperative \cite{2}. 

Central to the TI’s ambitious goal is its stringent network performance requirements. For mission-critical applications, TI necessitates a network latency typically between 1-10 milliseconds and a remarkably high packet delivery ratio of 99.99999\%. These specifications are vital due to the sensitivity of human touch and the potentially catastrophic outcomes of any failure in these systems \cite{3}.

TI's applications, notably in domains like remote surgery, demand ultra-low latency and high reliability and security. The variability in latency requirements, often less than 10 milliseconds, is dictated by the specific nature and dynamicity of the application. TI aspires to achieve an ultra-low end-to-end round-trip latency of 1 millisecond, setting a new benchmark in network performance \cite{4}.

Diverging from traditional robotic surgery, TI enables remote robotic surgery where a surgeon operates on a distant patient through a network. This requires unprecedented transparency to ensure that the surgeon’s actions are accurately mirrored in the patient-side domain and that the surgeon is precisely aware of the robotic arm's position at the patient side. Achieving this bidirectional awareness in the face of communication-induced delays is a significant technical challenge \cite{5}\cite{6}.

Moreover, the current 5G mobile networks only partially meet these stringent requirements of TI. Issues such as delay, packet loss, and jitter can critically impact the stability and safety of remote robotic surgery systems \cite{7}\cite{8}. Due to the highly time-sensitive requirements of the application domain, it is essential to explore computationally lightweight solutions. This paper introduces the application of Kalman Filter--assisted by an offline System Identification learning module--as a solution to these challenges. By accurately estimating the position of the PSM arm, the KF enhances the reliability and precision of TI applications, particularly in the high-stakes realm of remote surgery.

This paper extends our previous work presented in a conference paper, which employed the MATLAB System Identification Tool \cite{9}. In the current study, we use the MOESP method due to its superior ability to capture relationships within the data and generate a state-space model. This model is then utilized with the KF, significantly enhancing the precision of position estimation for the arm at the PSM side. This improvement is critical for the accuracy and reliability of remote robotic surgeries facilitated by the TI. The enhanced estimation precision provided by our approach aims to advance the effectiveness and safety of TI-enabled remote surgical procedures.

\section{RELATED WORK}
Remote robotic surgery facilitated by the TI requires highly accurate and real-time estimation of the position of surgical tools. However, challenges such as packet loss, network delays, and jitter can significantly impact the precision and reliability of these estimations. Previous approaches have attempted to address these challenges using various techniques, each with unique strengths and limitations.

In \cite{91}, the authors address packet loss and delay challenges in remote robotic surgery within a 5G Tactile Internet environment, advocating for a Gaussian process regression (GPR) approach to predict and compensate for delayed/lost messages. Two kernel versions of the sequential randomized low-rank and sparse matrix factorization method (1-SRLSMF and SRLSMF) were introduced to scale GPR for handling delayed/lost data in training datasets.This approach effectively models uncertainties and compensates for missing data, making it relevant to the accurate positioning of robotic arms during surgery. However, this approach faces challenges due to the computational complexity of Gaussian processes, especially the kernel matrix inversion, which escalates with increasing data points. 

In \cite{10}, the authors proposed a method based on deep learning and Convolutional Neural Networks (CNN) for evaluating surgical skills in robot-assisted surgery. It introduces a deep learning framework to assess skills by mapping motion kinematics data to skill levels using a Deep CNN. The study also highlights the limitations of CNN, including the need for improved labeling methods, optimization of the deep architecture, and exploring ways to visualize deep hierarchical representations to uncover hidden skill patterns. 

In \cite{29}, the authors proposed a human collective intelligence-inspired, multi-view representation learning approach. This method introduces view communication by simulating human communication mechanisms, enabling each view to exploit complementary information from other views to help model its representation. This approach has significantly improved classification accuracy across various fields, highlighting the importance of communication and information exchange in multi-view learning frameworks.

The data for this study is collected through the da Vinci Research Kit (dVRK), a specialized set of robotic tools for testing surgical procedures. It is the first generation model of the da Vinci Surgical System (dVSS) by a company called Intuitive Surgical \cite{11}. Experts have thoroughly studied this kit to understand system dynamics \cite{12} \cite{13}. This research delves into the PSM arm movements, using the JIGSAW data of the surgical system to learn its mechanics.

The main objective of this study is to explore the efficacy of the KF to estimate the position of the PSM's arm, even when the network experiences delays, jitter, or data packet loss. These issues are a significant challenge, especially when bidirectional touch information and precise control are necessary across the network. We conduct comprehensive simulations under various network conditions to address these challenges, rigorously testing the proposed system's performance. The simulations aim to mimic real-world scenarios and demonstrate the robustness and reliability of our approach, ultimately contributing to safer and more effective remote robotic surgeries.
\begin{figure}[H]
    \centering
    \includegraphics[width=0.5\textwidth]{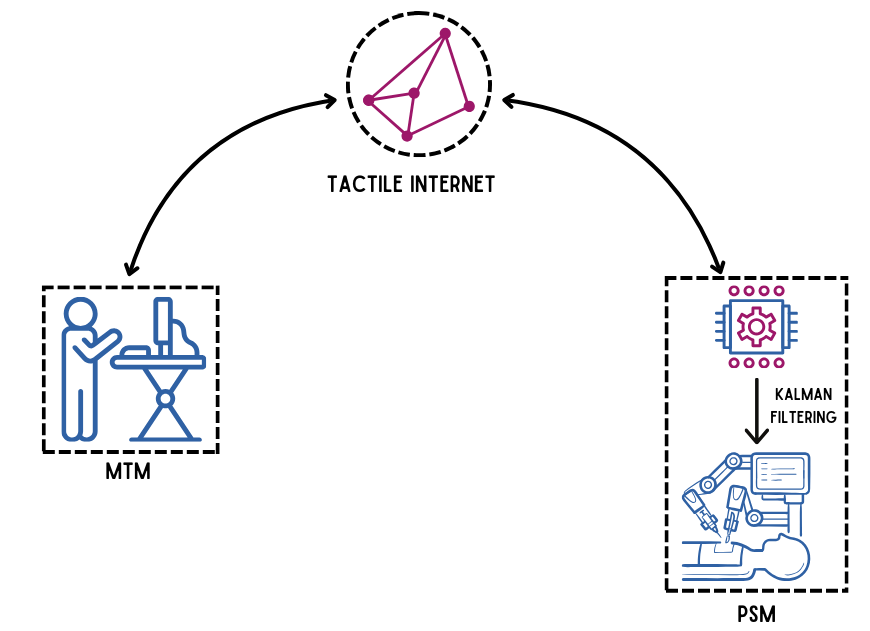}
    \caption{Remote Robotic Surgery Framework Utilizing TI and KF for Enhanced PSM Precision}
    \label{model}
\end{figure}

\section{METHODOLOGY}
The KF is a well-known signal processing algorithm that employs efficient, recursive computation for process state estimation, aiming to minimize the mean squared error. It supports estimating past, present, and future states, even amidst system uncertainties. It was introduced by Rudolf E. Kalman in 1960 \cite{116}, and it has been effectively used in fields requiring accurate and real-time estimation, such as navigation. It is particularly effective in systems where data is uncertain or noisy, which is common in remote robotic surgery due to sampling noise and network uncertainties.

\subsection{System Model}
The proposed architecture \cite{91}, as shown in Fig.~\ref{model}, is a remote robotic surgery system facilitated by the TI. The system is compartmentalized into three primary domains: the surgeon-side domain, the patient-side domain, and the network domain, each of which plays an integral role in the surgical procedure's execution.

\textbf{Surgeon-Side Domain}:
The surgeon-side domain comprises an ergonomically designed surgeon console/master tool manipulator (MTM) and the operating surgeon. The surgeon interacts with the console, capturing the surgeon's gestures and translating them into haptic commands. These commands encapsulate the surgeon's intended surgical maneuvers, encompassing force, orientation, and kinematic parameters.

\textbf{Patient-Side Domain}:
The patient-side domain hosts the PSM and the patient. Upon reception of the haptic commands, the PSM, equipped with an estimation KF algorithm (in our case KF), interprets these inputs to estimate and enact the precise movements corresponding to the surgeon's inputs. The KF algorithm is pivotal for real-time estimation and correction of the robot's arm position, as it assists in maintaining the fidelity of the surgical gestures amidst potential perturbations in signal transmission.

\textbf{Network Domain}:
Central to the communication bridge between the surgeon and patient domains, the network domain is tasked with delivering low-latency and ultra-reliable connectivity. 

\begin{figure*}[htbp]
    \centering
    \includegraphics[width=\textwidth]{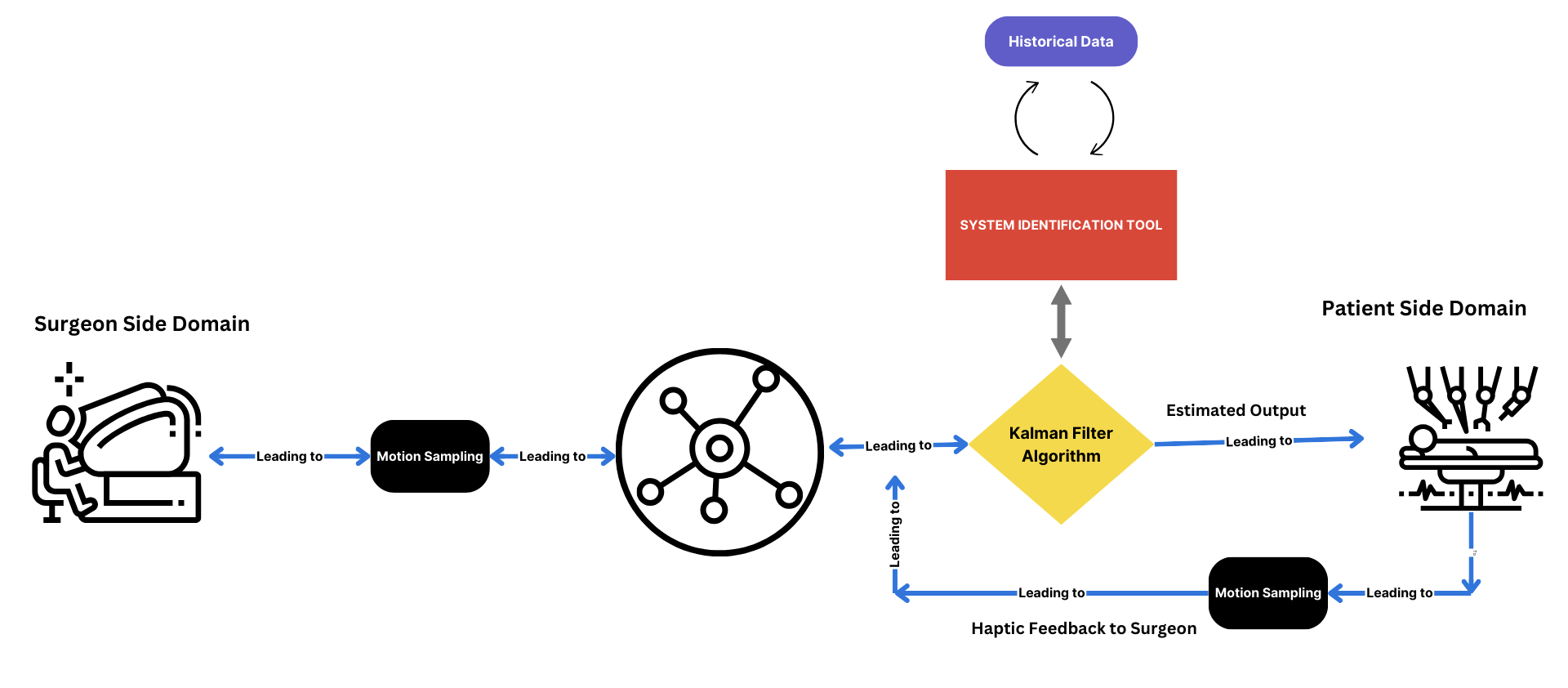}
    \caption{Operational Sequence of Remote Surgery Using TI and KF Algorithm}
    \label{fig:model1}
\end{figure*}

\textbf{Operational Workflow}:
The block diagram, as shown in Fig.~\ref{fig:model1}, illustrates the operational sequence that initiates with the surgeon interacting with the MTM, generating a set of haptic commands. These commands are transmitted through the network domain, leveraging TI technology. The movements are processed using a KF Algorithm, which utilizes historical data for system identification and accurately estimates the required movement. The PSM then executes this estimated output in the patient-side domain. Feedback from the PSM is returned to the surgeon, providing vital tactile information to inform the surgeon's subsequent movements. This feedback loop is essential for the precision and safety of remote surgical procedures.

\subsection{KF Implementation for State Estimation with TI Network Effects}
For completeness, we briefly describe the KF fundamentals in this section. The KF has numerous applications in technology. A typical application is the guidance, navigation, and control of vehicles, especially aircraft and exploration robots \cite{116}. It is also widely used in signal processing and Quantum Systems \cite{216}.

The system is represented as follows:

\begin{itemize}
    \item \textbf{State Equation (System Dynamics):} The true state of the system evolves according to the discrete-time state-space model. Still, it is affected by the network characteristics before being observed at the PSM side.
    \begin{equation}
        \tilde{x}_k = A\tilde{x}_{k-1} + Bu_{k-1} + w_{k-1}
    \end{equation}
    where:
    \begin{itemize}
        \item \( \tilde{x}_k \) is the predicted state vector of the system at time \( k \) before accounting for network effects.
        \item \( A \) is the state transition matrix that models the system dynamics from one timestep to the next.
        \item \( B \) is the control input matrix that translates the input commands into changes in the state.
        \item \( u_{k-1} \) is the control input vector applied at the previous time step \( k-1 \).
        \item \( w_{k-1} \) represents the process noise at the previous time step, encompassing the system model's inherent uncertainties and additional perturbations such as those introduced by the network jitter and packet loss.
    \end{itemize}
    
    \item \textbf{Network Effects:} The network introduces additional deviations to the state vector as transmitted from the MTM to the PSM. These deviations are modeled as:
    \begin{equation}
        x_k = \tilde{x}_k + n_{d} + n_{j} + n_{p}
    \end{equation}
    where:
    \begin{itemize}
        \item \( x_k \) is the state vector as it arrives at the PSM, having been affected by the network.
        \item \( n_{d} \) models the deviation caused by network delay, which can vary depending on network conditions.
        \item \( n_{j} \) models the deviation caused by network jitter, representing the variability in the delay.
        \item \( n_{p} \) models the deviation caused by packet loss, which can result in intermittent information losses.
    \end{itemize}
    
    \item \textbf{Measurement Equation (Observation Model):} The PSM observes the system state with its sensors, which the measurement equation can represent.
    \begin{equation}
        z_k = Hx_k + v_k
    \end{equation}
    where:
    \begin{itemize}
        \item \( z_k \) is the measurement vector at time \( k \).
        \item \( H \) is the observation matrix that relates the state vector to the measurements.
        \item \( v_k \) represents the measurement noise at time \( k \), reflecting the sensor noise and other observational inaccuracies.
    \end{itemize}
\end{itemize}

Given these models, the KF operates in two steps, Prediction and Update, to estimate the system state:
\begin{itemize}
    \item \textbf{Prediction Step:} The filter predicts the state of the system at the next time step along with the estimation uncertainty.
    \begin{align}
        \hat{x}_k^- &= A\hat{x}_{k-1} + Bu_{k-1} \\
        P_k^- &= AP_{k-1}A^T + Q
    \end{align}
    where:
    \begin{itemize}
        \item \( \hat{x}_k^- \) is the a priori estimate of the state vector before the measurement at time \( k \) is taken into account.
        \item \( P_k^- \) is the a priori estimate of the state covariance, indicating the uncertainty of the prediction.
        \item \( Q \) is the covariance matrix of the process noise, quantifying the expected variance in the predictions due to the inherent uncertainty in the system dynamics and the effect of the network.
    \end{itemize}
    
\item \textbf{Update Step:} The filter then incorporates the new measurement to refine its state vector estimate and update the estimation uncertainty.
    \begin{align}
        K_k &= P_k^- H^T (HP_k^- H^T + R)^{-1} \\
        \hat{x}_k &= \hat{x}_k^- + K_k(z_k - H\hat{x}_k^-) \\
        P_k &= (I - K_kH)P_k^-
    \end{align}
    where:
    \begin{itemize}
        \item \( K_k \) is the Kalman gain at time \( k \), which determines how much the predictions should be adjusted based on the new measurement.
        \item \( \hat{x}_k \) is the a posteriori estimate of the state vector after incorporating the measurement at time \( k \).
        \item \( P_k \) is the a posteriori estimate of the state covariance, indicating the updated uncertainty of the state estimate.
        \item \( R \) is the covariance matrix of the measurement noise, quantifying the expected variance in the measurements.
        \item \( I \) is the identity matrix, with the same dimensions as \( P_k^- \).
    \end{itemize}
\end{itemize}

The KF uses these equations to continuously estimate the system's state in the presence of noise and uncertainties, including those introduced by the TI. The filter's ability to assess the actual state in such an environment measures its robustness and effectiveness.

\subsection{Empirical Estimation of Q and R}
In Kalman Filtering, the performance of the filter is heavily influenced by the accurate specification of the process noise covariance matrix ($\mathbf{Q}$) and the measurement noise covariance matrix ($\mathbf{R}$). These matrices are derived from prior knowledge of the system's noise characteristics. However, in practical scenarios where this information is unavailable, a data-driven approach can be used to empirically estimate $\mathbf{Q}$ and $\mathbf{R}$ from the available data.
%\subsubsection{Empirical Estimation Approach}
The empirical estimation method involves the following steps:

1. \textbf{Initialization}: Begin with initial guesses for the covariance matrices $\mathbf{Q}$ and $\mathbf{R}$. These initial guesses can be small positive definite matrices, reflecting low uncertainty.
\begin{equation}
\mathbf{Q}_{\text{initial}} = \epsilon_Q \cdot \mathbf{I}, \quad \mathbf{R}_{\text{initial}} = \epsilon_R \cdot \mathbf{I}
\end{equation}

where $\epsilon_Q$ and $\epsilon_R$ are small positive constants, and $\mathbf{I}$ is the identity matrix.

2. \textbf{Initial Kalman Filter Run}: Perform an initial run of the Kalman Filter using these initial guesses. This run provides preliminary state and measurement estimates.

3. \textbf{Calculation of Residuals}: 
   \begin{itemize}
   \item \textbf{Measurement Residuals} ($r_y$): These are the differences between the actual measurements ($y$) and the estimated measurements ($\hat{y}$).

\begin{equation}
r_y(k) = y(k) - \hat{y}(k)
\end{equation}

   \item \textbf{Process Residuals} ($r_x$): These are the differences between consecutive state estimates ($\hat{x}$).

\begin{equation}
    r_x(k) = \hat{x}(k) - \mathbf{A} \hat{x}(k-1) - \mathbf{B} u(k)
\end{equation}

   \end{itemize}

4. \textbf{Empirical Covariance Calculation}: 
   \begin{itemize}
   \item The empirical measurement noise covariance matrix ($\mathbf{R}_{\text{empirical}}$) is calculated as the covariance of the measurement residuals:
\begin{equation}
    \mathbf{R}_{\text{empirical}} = \frac{1}{N} \sum_{k=1}^{N} r_y(k) r_y(k)^T
\end{equation}

   \item The empirical process noise covariance matrix ($\mathbf{Q}_{\text{empirical}}$) is calculated as the covariance of the process residuals:
\begin{equation}
\mathbf{Q}_{\text{empirical}} = \frac{1}{N-1} \sum_{k=1}^{N-1} r_x(k) r_x(k)^T
\end{equation}

   \end{itemize}

The empirically estimated $\mathbf{Q}$ and $\mathbf{R}$ are then used in subsequent runs of the KF to improve its performance.

%\section{Dataset and Experimental Setup}
\section{SOURCE AND COMPARISON OF THE JIGSAW DATASET}
\begin{figure*}[htbp]
  \centering
  \includegraphics[width=\textwidth]{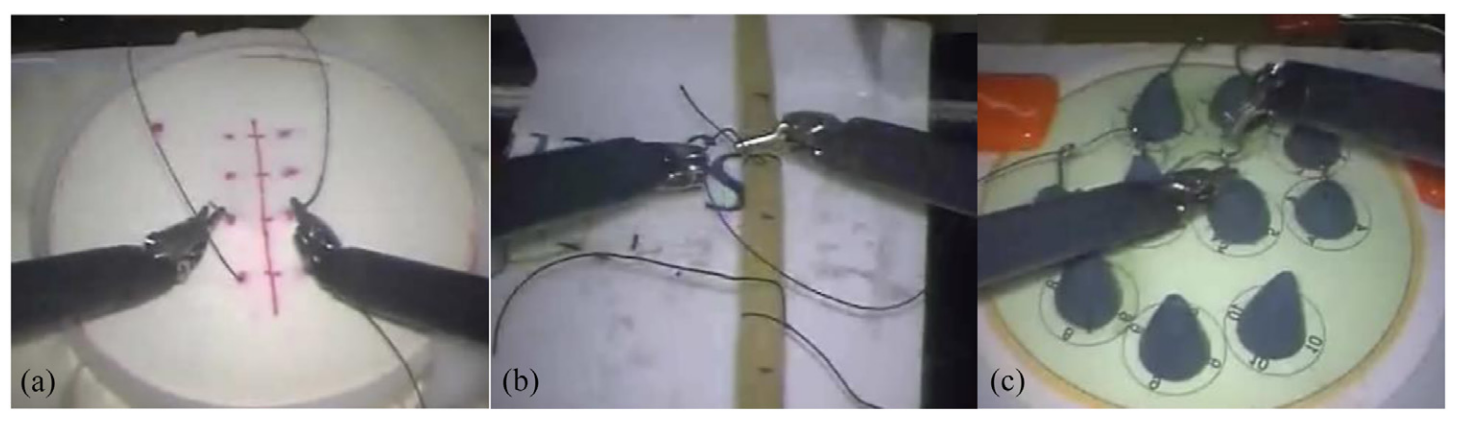}
  \caption{Three surgical tasks: (a) Suturing, (b) Knot-tying and (c) Needle-passing  \cite{21}.}
  \label{fig:data}
\end{figure*}

To evaluate the remote robotic surgery system, we used the JIGSAWS dataset, a comprehensive surgical skill dataset developed by the Computational Interaction and Robotics Laboratory at Johns Hopkins University \cite{17}. JIGSAWS stands for the JHU-ISI Gesture and Skill Assessment Working Set and encompasses kinematic, video, and gesture data from three elementary surgical tasks performed using the da Vinci surgical robot: suturing, knot tying, and needle passing as shown in Fig.~\ref{fig:data}.

\begin{figure}[ht]
\centering
\includegraphics[width=3.75in]{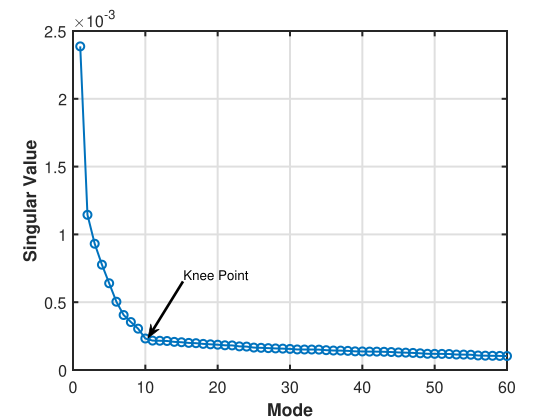}
    \caption{Singular Values from MOESP Algorithm}
    \label{fig:SVD}
\end{figure}

The dataset captures the motion data of both the master controls and the corresponding slave manipulators (end-effectors) during the execution of the tasks. A diverse group of eight participants, ranging from novices to experts, contributed to the dataset. Each participant performed each surgical task five times, resulting in 40 trials. The kinematic data in JIGSAWS captures the Cartesian positions \((\mathbf{p} \in \mathbb{R}^3)\), rotation matrices \((\mathbf{R} \in \mathbb{R}^{3 \times 3})\), linear velocities \((\mathbf{v} \in \mathbb{R}^3)\), rotational velocities \((\boldsymbol{\omega} \in \mathbb{R}^3)\), and grasper angles \((\theta)\) for left and right tools for both PSM and MTM, resulting in a total of 76 features sampled at 30Hz. The video data is also collected at 30fps from an endoscopic camera \cite{18}.

\section{SUBSPACE IDENTIFICATION USING MOESP}
\subsection{Introduction to Subspace Identification}
Subspace identification methods are a family of algorithms used to identify state-space models of dynamic systems from input-output data. These methods are particularly valuable in situations where the underlying state-space structure of the system is complex and not easily modeled with classical parametric techniques. Subspace methods operate by decomposing the data into subspaces that capture the most informative dynamics of the system. This approach is advantageous for its robustness to noise and its ability to handle large datasets efficiently. It is ideal for applications in remote robotic surgery where accuracy and computational efficiency are crucial. While implementing the KF in the da Vinci Surgical System (dVSS), accurately modeling system dynamics and noise characteristics is paramount. However, the proprietary nature of dVSS poses a challenge as its dynamics and noise characteristics are not publicly disclosed \cite{7}.

In \cite{14}, authors present a Convex Optimization-based Dynamic Model Identification Package for the da Vinci Research Kit (dVRK), a teleoperated surgical robotic system. The package is designed to model the mechanical components of the dVRK and identify dynamic parameters subject to physical consistency. It addresses the need for accurate dynamic models before implementing robust model-based control algorithms and is open-source, making it feasible for similar robots.

In \cite{19}, authors describe the development of a dynamic simulator for the dVRK (PSM) in the CoppeliaSim robotic simulation environment. The simulator aims to accurately predict the behavior of the real robot by integrating the kinematic and dynamic properties, including the double parallelogram and the counterweight mechanism.

The JIGSAWS dataset poses a significant challenge due to its high-dimensional nature, featuring 76 inputs that capture the complex motions of the MTM and PSM. The inherent nonlinearity in the kinematic behavior of the robotic system, especially in the PSM arm's x, y, and z positional data, neceto estimateted modeling approach capable of efficiently handling nonlinear dynamics.

To address these challenges, this study adopts the MOESP method, inspired from the subspace identification method from \cite{b25}. Unlike traditional approaches such as the nonlinear Auto-regressive with exogenous input (ARX) model used previously\cite{9}, the MOESP method offers a more coherent approach to system identification, particularly when dealing with high-dimensional data.The code for the MOESP system identification used in this study was taken from \cite{b26}.

The MOESP method is one of the vital subspace identification techniques. It excels explicitly in environments where the data involve multiple inputs and outputs, which is typical in complex systems like those of the robotic manipulators used in surgery. MOESP constructs state-space models by identifying subspaces from the data that correspond to the system's dynamics. It provides the state space model of the dVRK robotic system from the JIGSAW data set,  which we have used in KF to estimate the position of the arm at PSM side.

Incorporating the MOESP method for system identification in remote robotic surgery systems significantly enhances the estimation accuracy. This, in turn, improves the effectiveness of the KF, resulting in better precision and reliability of surgical operations facilitated by the TI.

\subsection{DATA ORGANIZATION AND MATRIX CONSTRUCTION}
Using the MOESP method, data from the robotic surgery system is organized into Hankel matrices, efficiently capturing the system's dynamics across multiple time steps.

The detailed equations are mentioned in \cite{b25} regarding MOESP and N4SID. In our study, we consider the Hankel matrix, which is constructed from the output data, and plays a crucial role in the MOESP method.

\begin{equation}
    H_y = \begin{bmatrix}
    y(1) & y(2) & \cdots & y(n) \\
    y(2) & y(3) & \cdots & y(n+1) \\
    \vdots & \vdots & \ddots & \vdots \\
    y(m) & y(m+1) & \cdots & y(m+n-1)
    \end{bmatrix}
\end{equation}
where \(y(t)\) represents the output at time \(t\), and \(m\) and \(n\) define the matrix dimensions, capturing the evolving dynamics of the robotic arm movements.

\subsubsection{Importance of Hankel matrices}
Hankel matrices are integral for MOESP as they enable the capturing and structuring of system dynamics which are essential for accurate state-space modeling.
Hankel matrices are used in MOESP to organize input-output data into a form that captures the inherent dynamics of the surgical system across time. This organization is critical for effectively extracting system characteristics and subsequent control actions.

Following are the step-by-step description of the MOESP method used in our work. 
\begin{itemize}
    \item Structuring the data to capture temporal dynamics effectively.
    \item Facilitating the mathematical operations needed for subspace identification, such as projections and decompositions.
    \item Enhancing the precision of the identified model by ensuring that the data encompasses sufficient system behavior over time.
\end{itemize}
Followings is the procedure that indicates a structured method for identifying the state-space model. 
\subsubsection{Identification Method for State-Space Model Estimation}

\textbf{Step 1: Matrix Construction}\\
Construct Hankel matrices for both input \(U\) and output \(Y\) data, encompassing both past and future information:
\begin{equation}
    W_p = \begin{bmatrix} U_p \\ Y_p \end{bmatrix}, \quad W_f = \begin{bmatrix} U_f \\ Y_f \end{bmatrix}
\end{equation}

\textbf{Step 2: Subspace Decomposition}\\
Project the future output matrix \(Y_f\) onto the subspace spanned by the past data \(W_p\) using the oblique projection:
\begin{equation}
    Y_f = \Theta_k X_f + \Psi_k U_f,
\end{equation}
where \(\Theta_k\) is the extended observability matrix derived from \(W_p\).

\textbf{Step 3: QR and SVD Decompositions}\\
Apply QR decomposition to \(W_p\) to facilitate the decomposition of \(Y_f\), followed by Singular Value Decomposition (SVD) to isolate significant state dynamics:
\begin{equation}
    \text{SVD}(\Theta_k X_f) \rightarrow \text{system order and dynamics}.
\end{equation}

\textbf{Step 4: State-Space Model Estimation}\\
Estimate the matrices \(A\), \(B\), \(C\), and \(D\) that describe the dynamics of the robotic system. This estimation is critical for accurately predicting and controlling the robotic arm's movements during surgery.

 Fig.~\ref{fig:SVD} shows the singular values obtained from applying the MOESP algorithm to the normalized input and output data. The singular values, plotted in descending order, help identify the significant modes of the system. The knee point around mode 10 indicates the practical order of the system. Modes beyond this point primarily capture noise rather than meaningful system dynamics.

\subsubsection{\textbf{Validation of Model Predictions}}
To validate the MOESP model, we used a cross-validation methodology. We employed an independent dataset, different from the one used for system identification, to evaluate the model's predictive performance. Our main goal was to assess how well the model could predict the PSM arm's x, y, and z positions. These positions are crucial for accurately translating haptic commands during real-time surgery.

Additionally, this paper presents an efficient algorithm that uses a KF approach to address network-related challenges in remote robotic surgery. It accurately estimates the PSM position, considering network delay, jitter, and packet loss, as detailed in Algorithm 1.
Moreover, the state-space matrices $A$, $B$, $C$, and $D$ are derived from MOESP, representing the identified system. The detailed steps of the MOESP algorithm are outlined in the Algorithm 2.

\begin{algorithm}
\caption{KF with Network Effects for State Estimation}
\begin{algorithmic}[1]

\STATE \textbf{Input:} MTM, PSM, A, B, C, D, Q, R
\STATE \textbf{Output:} z\_est, MSE, Est\%

\STATE Init: $x\_est, P, prev\_y, N, dt, nd, nj, np$
\FOR {$k = 2$ to $N$ Samples}
    \STATE $del\_k \gets \max(1, k - \text{round}(nd / dt + \text{randn}() \times nj / dt))$
    \STATE $y \gets (\text{rand}() > np) ? PSM[del\_k] : prev\_y$
    \STATE $prev\_y \gets y$, $x\_pred \gets A x\_est[k-1] + B MTM[k-1]$
    \STATE $P\_pred \gets A P A^T + Q$
    \FOR {$d = 1$ to $\text{size}(C, 1)$}
        \STATE $K \gets P\_pred C[d]^T / (C[d] P\_pred C[d]^T + R[d])$
        \STATE $x\_est[k] \gets x\_pred + K (y[d] - C[d] x\_pred)$
        \STATE $P \gets (I - K C[d]) P\_pred$
    \ENDFOR
\ENDFOR
\STATE $z\_est \gets C x\_est$, Calc RMSE, Est\%

\end{algorithmic}
\end{algorithm}

\begin{algorithm}
\caption{MOESP for State-Space System Identification}
\begin{algorithmic}[1]
\STATE \textbf{Input:} Normalized input data $U_{\text{norm}}$, normalized output data $V_{\text{norm}}$, number of samples $N$, block size $d$
\STATE \textbf{Output:} Singular values $ss$, state-space matrices $A$, $B$, $C$, $D$
\STATE \textbf{Initialization:}
\STATE $N_{\text{samples}} \gets 1240$ \COMMENT{Number of samples}
\STATE $N_{\text{vars}} \gets 3$ \COMMENT{Number of variables}
\STATE $U_{\text{norm}} \gets \text{Min-Max normalized MTM data}$ \COMMENT{Normalized input data}
\STATE $V_{\text{norm}} \gets \text{Min-Max normalized PSM data}$ \COMMENT{Normalized output data}
\STATE Form block Hankel matrices $Y$ and $U$ using $d$ and $N$
\FOR {$s = 1$ to $d$}
    \STATE $Y((s-1)N_{\text{vars}}+1:sN_{\text{vars}}, :) \gets V_{\text{norm}}(s:s+N-1, :)$
    \STATE $U((s-1)N_{\text{vars}}+1:sN_{\text{vars}}, :) \gets U_{\text{norm}}(s:s+N-1, :)$
\ENDFOR
\STATE Perform LQ decomposition on $[U; Y]^T$
\STATE $R \gets \text{triu}(\text{qr}([U; Y]^T)^T)$
\STATE Extract $R_{22} \gets R(d \times N_{\text{vars}}+1:end, d \times N_{\text{vars}}+1:end)$
\STATE Perform SVD on $R_{22}$
\STATE $[U_1, S_1] \gets \text{svd}(R_{22})$
\STATE Extract singular values $ss \gets \text{diag}(S_1)$
\STATE \textbf{System Order Selection:} Choose $n$ based on singular values (e.g., 85\% energy criterion)
\STATE \textbf{State-Space Matrices Calculation:}
\STATE $O_k \gets U_1(:, 1:n) \times \text{diag}(\sqrt{ss(1:n)})$
\STATE $C \gets O_k(1:N_{\text{vars}}, :)$
\STATE $A \gets O_k(1:N_{\text{vars}} \times (d-1), :) \backslash O_k(N_{\text{vars}}+1:d \times N_{\text{vars}}, :)$
\STATE $L_1 \gets U_1(:, n+1:end)^T$
\STATE $R_{11} \gets R(1:d \times N_{\text{vars}}, 1:d \times N_{\text{vars}})$
\STATE $R_{21} \gets R(d \times N_{\text{vars}}+1:end, 1:d \times N_{\text{vars}})$
\STATE $M_1 \gets L_1 \times R_{21} \times R_{11}^{-1}$
\STATE Form $M$ and $L$ matrices for the solution
\STATE $DB \gets L \backslash M$
\STATE Extract $D \gets DB(1:N_{\text{vars}}, :)$
\STATE Extract $B \gets DB(N_{\text{vars}}+1:end, :)$
\end{algorithmic}
\end{algorithm}

\subsection{Data Preprocessing}
We employed Min-Max Normalization to scale the input data to preprocess the JIGSAWS dataset. This technique typically transforms features by scaling them to a given range [0, 1]. The formula used for Min-Max Normalization is:
\[
X' = \frac{X - X_{\min}}{X_{\max} - X_{\min}}
\]
where \( X \) is the original feature vector, \( X_{\min} \) and \( X_{\max} \) are the minimum and maximum values of the feature, respectively. This normalization ensures that all features contribute equally to the model, preventing features with more extensive ranges from dominating those with smaller ranges. By normalizing the data, we improve the stability and performance of the Kalman Filter, especially in handling the high-dimensional and diverse inputs from the JIGSAWS dataset.

\section{RESULTS}
The implementation of the KF for state estimation through a network characterized by delay, jitter, and packet loss was evaluated. The network parameters were set to simulate a best-effort network scenario, reflecting conditions that might commonly be encountered in real-world Tactile Internet applications.
The network simulation parameters were as follows:
\begin{itemize}
    \item Network delay (\( n_d \) in ms): This represents a constant time delay that every packet experiences during transmission over the network. In \cite{22}, the TI delay range is mentioned for 5G services and use cases.
    \item Jitter variance (\( n_j \) ms):  This is the variance of the jitter, indicating the degree of random fluctuation in the timing of packet arrivals around the mean network delay. In \cite{23}, authors discuss the impact of jitter in 5G networks on the performance of real-time services. Values of less than 0.01 seconds are associated with good performance. 
    \item Packet loss probability (\( n_p \) in \%): This is the likelihood that any given packet will be lost during transmission and not reach its destination. Values between 0.01 and 0.1 (1\% to 10\%) are often used in simulations to study the impact of packet loss \cite{24}.
\end{itemize}
Under these conditions, the KF estimated the PSM's state using MTM inputs. The estimation's effectiveness was measured by the percentage accuracy, reflecting the similarity between the estimated and acPSM states. The reference values for network conditions such as jitter variance, network delay, and packet loss probability were obtained from Table \ref{tab:reliability_latency_requirements} from \cite{28} to ensure our simulations were based on established benchmarks.

\subsection{Summary of Position x, y, z Graphs}
The KF estimation across the three dimensions—Position X, Y, and Z—shows reasonably good performance, though with some variability in accuracy and RMSE values.

For Position X, the RMSE is 0.0331 with an accuracy of 94.80\%. While this indicates a decent alignment between the KF estimates and the ground truth, the estimation isn't flawless, especially in areas where rapid changes occur. The KF generally tracks the data well but may struggle with precise tracking during more complex variations.

Position Y displays a slightly better RMSE of 0.0297 but a lower accuracy at 95.99\%. The KF's performance here suggests it is reasonably reliable, though slight deviations from the ground truth can be observed, particularly in more intricate parts of the trajectory. The overall trend is captured, but some finer details may be lost.

Position Z, with an RMSE of 0.0243 and the highest accuracy of 97.67\%, seems to be where the KF performs best. However, even in this dimension, the KF is not immune to occasional mismatches with the ground truth, though these are less frequent than in the other dimensions.

Overall, the KF provides robust estimations with accuracies mostly above 95\%, making it a viable tool for scenarios requiring positional tracking. However, there are areas, especially during rapid or complex changes, where its performance could be improved to achieve closer alignment with the ground truth.
%The KF estimation for Position x, y, and z demonstrates high accuracy and low RMSE values under different conditions. For Position x in Fig.~\ref{fig:Outputx}, the RMSE is 0.0331 with an accuracy of 96.80\%, indicating a good fit between the KF estimate and the ground truth and effective tracking of variations. Position y in Fig. \ref{fig:Outputy} shows a lower RMSE of 0.0297 with an accuracy of 95.99\%. Here, the KF closely follows the ground truth data, providing reliable estimations. Position z in Fig. \ref{fig:Outputz} also has an RMSE of 0.0243 but achieves the highest accuracy at 97.67\%, suggesting that the KF is particularly effective at estimating variations in Position z. Overall, the KF maintains accuracies above 95\% for all positions, demonstrating robustness and reliability and thus making it suitable for tactile internet applications requiring precise positional data.%}
The detailed scenarios are as detailed in Table. \ref{tab:network_params}. 
%\begin{figure}[htbp]
%    \centering   \includegraphics[width=\textwidth]{Position_x.eps}
 %   \caption{Comparison of PSM data with the KF estimated output for Position x}
  %  \label{fig:Outputx}
%\end{figure}
%\begin{figure}[htbp]
   % \centering    \includegraphics[width=\textwidth]{Position_y.eps}
    %\caption{Comparison of PSM data with the KF estimated output for Position y}
    %\label{fig:Outputy}
%\end{figure}
%\begin{figure}[htbp]
    %\centering  \includegraphics[width=\textwidth]{Position_z.eps}
    %\caption{Comparison of PSM data with the KF estimated output for Position z}
    %\label{fig:Outputz}
%\end{figure}

\begin{figure}[ht]
\centering
\includegraphics[width=3.75in]{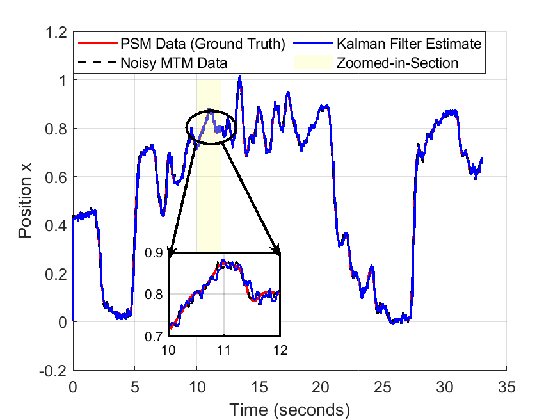} 
    \caption{Comparison of PSM data with the KF estimated output for Position x}
    \label{fig:Outputx}
\end{figure}

\begin{figure}[ht]
\centering
\includegraphics[width=3.75in]{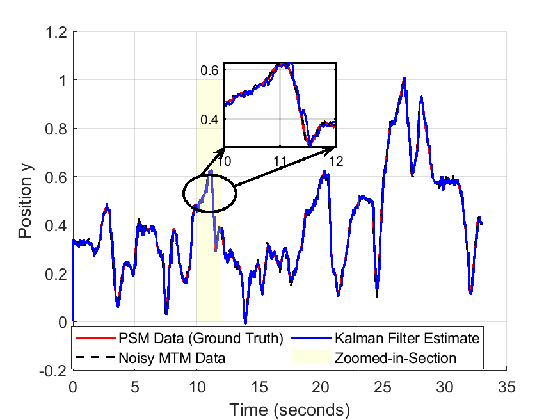} 
    \caption{Comparison of PSM data with the KF estimated output for Position y}
    \label{fig:Outputy}
\end{figure}
\begin{figure}[ht]
\centering
\includegraphics[width=3.75in]{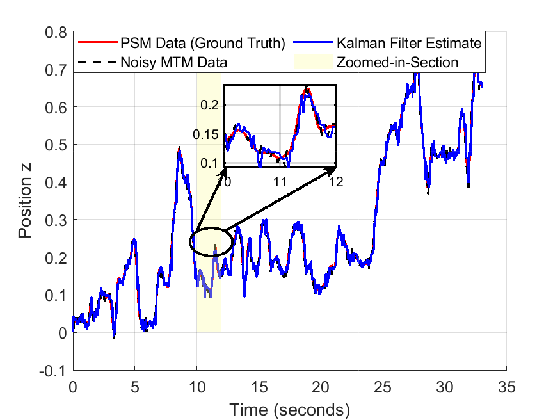} 
    \caption{Comparison of PSM data with the KF estimated output for Position z}
    \label{fig:Outputz}
\end{figure}

\begin{table*}[htbp]
    \centering
    \caption{Reliability and Latency Requirements for Selected Tactile Internet Applications}
    \begin{tabular}{|c|c|c|c|}
        \hline
        \textbf{Applications} & \textbf{Cases} & \textbf{Delay (ms)} & \textbf{Reliability \%} \\
        \hline
        Haptics & Highly-dynamic: 0.5, Dynamic: 5, Fixed: 50 & 0.5 – 2 & 99.9 \\
        \hline
        Augmented Reality & Video & 0.5 – 2 & \> 99.999 \\
        \hline
        Immersive Reality & 3D-Audio & 0.5 – 2 & 99.9 \\
        \hline
    \end{tabular}%
    \label{tab:reliability_latency_requirements}
\end{table*}
\begin{table*}[htbp]
    \centering
    \caption{Network Parameters and KF Estimation Accuracy for Tactile Internet Scenarios}
   \renewcommand{\arraystretch}{}
    \resizebox{\textwidth}{!}{%
    \begin{tabular}{|c|c|c|c|c|c|c|c|c|}
        \hline
        \textbf{Jitter Variance (ms)} & \textbf{Network Delay (ms)} & \textbf{Packet Loss Probability (\%)} & \multicolumn{3}{c|}{\textbf{KF Est. Accuracy (\%)}} & \multicolumn{3}{c|}{\textbf{KF RMSE}} \\
        \cline{4-9}
         &  &  & \textbf{Position x} & \textbf{Position y} & \textbf{Position z} & \textbf{Position x} & \textbf{Position y} & \textbf{Position z} \\
        \hline
        0.5 & 0.5 - 2 & 0.01 & 96.67 & 95.13 & 97.52 & 0.0363 & 0.0308 & 0.0333 \\
        \hline
        0.5 & 0.5 - 2 & 0.001 & 95.58 & 95.61 & 97.60 & 0.0356 & 0.0310 & 0.0337 \\
        \hline
        0.1 & 1 & 0.01 & 98.68 & 98.70 & 99.00 & 0.0232 & 0.0196 & 0.0239 \\
        \hline
        2.0 & 5 & 0.001 & 88.46 & 89.32 & 93.65 & 0.0600 & 0.0575 & 0.0581 \\
        \hline
        1.0 & 1 & 0.001 & 90.66 & 88.67 & 93.47 & 0.0644 & 0.0600 & 0.0523 \\
        \hline
        3.0 & 200 - 5000 & 1 & 85.95 & 87.68 & 90.95 & 0.0811 & 0.0766 & 0.0748 \\
        \hline
    \end{tabular}%
    }
    \label{tab:network_params}
\end{table*}

\section{ANALYSIS OF KF's ESTIMATION ACCURACY AND RSME UNDER VARIOUS NETWORK CONDITIONS}

\paragraph{Optimal Conditions:}
Under optimal network conditions with a jitter variance of 0.1 ms, network delay of 1 ms, and a packet loss probability of 0.01\%, the KF achieves its highest estimation accuracy. The accuracy rates for Positions x, y, and z are 98.68\%, 98.70\%, and 99.00\%, respectively, with shallow RMSE values of 0.0232, 0.0196, and 0.0239. This indicates that the KF performs exceptionally well in near-ideal conditions, making it highly suitable for applications requiring precise and accurate performance, such as remote robotic surgery and other critical tactile internet applications.

\paragraph{Mild Network Impairments:}
When mild network impairments are introduced, with a jitter variance of 0.5 ms, network delay ranging from 0.5 to 2 ms, and a packet loss probability of 0.01\%, there is a slight drop in the KF's accuracy. The estimation accuracy for Positions x, y, and z are 96.67\%, 95.13\%, and 97.52\%, respectively, and the RMSE values are 0.0363, 0.0308, and 0.0333. Despite this slight decrease in accuracy, the system maintains high reliability and accuracy, making it robust enough for applications that can tolerate minor network disruptions.

\paragraph{Moderate Network Impairments:}
In scenarios with moderate network impairments, where the jitter variance is 1.0 ms, network delay is 1 ms, and packet loss probability is 0.001\%, the KF's accuracy declines further. The accuracy rates for Positions x, y, and z drop to 90.66\%, 88.67\%, and 93.47\%, respectively, with corresponding RMSE values of 0.0644, 0.0600, and 0.0523. While the KF's performance is still acceptable for many applications, it may not be sufficient for those requiring the highest precision. This indicates the need for more robust handling of moderate network impairment cases.

\paragraph{High Network Impairments:}
High network impairments, characterized by a jitter variance of 2.0 ms, network delay of 5 ms, and packet loss probability of 0.001\%, result in a further reduction in the KF's accuracy. The KF estimation accuracies for Positions x, y, and z are 88.46\%, 89.32\%, and 93.65\%, respectively, with RMSE values of 0.0600, 0.0575, and 0.0581. These statistics reflect a noticeable drop in filter performance, particularly for Position x. Although the filter remains functional and can still be used for less precision-critical tasks, its effectiveness is significantly diminished in the presence of high network impairments.

\paragraph{Severe Network Impairments:}
Under severe network conditions, with a jitter variance of 3.0 ms, network delay ranging from 200 to 5000 ms, and a packet loss probability of 1\%, the KF's performance is at its lowest. The estimation accuracy drops to 85.95\% for Position x, 87.68\% for Position y, and 90.95\% for Position z, with RMSE values of 0.0811, 0.0766, and 0.0748. This significant reduction in accuracy highlights the challenges faced by the system in the face of extreme network impairments. The need for more robust algorithms and techniques to handle such severe impairments is evident, as it is crucial to maintain reliable performance even under highly adverse scenarios.

\section{Discussion}
The MOESP-based KF approach significantly reduces computational complexity compared to other state-of-the-art methods, making it exceptionally well-suited for real-time applications in Tactile Internet environments. Specifically, the KF's linear time complexity ($O(n)$) enables rapid processing, crucial for remote robotic surgery's high-frequency data and low-latency requirements. In contrast, Transformer models, while powerful, suffer from higher computational demands with complexities ranging from $O(n^2 \cdot d)$ to $O(n^3)$, which may introduce delays in real-time scenarios. Similarly, the Gaussian Process Regression (GPR) and Direct Robust Matrix Factorization (DRMF) framework, with a complexity of $O(n^3)$, further exacerbates these delays due to its intensive computational requirements, making it less practical for real-time operation. The comparison summarized in Table \ref{table:comparison} highlights these differences, underscoring the robustness, efficiency, and adaptability of our KF-based approach in handling the challenges inherent in Tactile Internet applications
\begin{table*}[h]
\centering
\caption{Comparison of Different Methods for Remote Robotic Surgery in Tactile Internet}
\label{table:comparison}
\begin{tabular}{|l|c|c|c|}
\hline
\textbf{Feature} & \textbf{MOESP-based KF} & \textbf{Transformer Models \cite{29}, \cite{31}} & \textbf{GPR \cite{91} and DRMF \cite{30}} \\ \hline
Computational Complexity & \textbf{Low ($O(n)$)} & Moderate to High ($O(n^2 \cdot d)$ to $O(n^3)$) & High ($O(n^3)$) \\ \hline
Real-Time Performance & High & Moderate & Limited \\ \hline
Handling Packet Loss & Robust & Moderate & Sensitive \\ \hline
Adaptability & High & Low & Limited \\ \hline
Data Requirements & Moderate & High & High \\ \hline
Scalability & Excellent & Moderate & Challenging \\ \hline
Robustness to Noise & Excellent & Moderate & Limited \\ \hline
Training Time & Minimal & Significant & Considerable \\ \hline
Reliability & High & Moderate & Moderate \\ \hline
Real-World Applications & Ideal for Tactile Internet & Limited & Theoretical Focus \\ \hline
\end{tabular}
\end{table*}

\section{CONCLUSION AND FUTURE RESEARCH DIRECTIONS}
This paper presents an advanced approach to position estimation in Tactile Internet-enabled remote robotic surgery, utilizing the KF enhanced by the MOESP method for state-space model identification. The KF is employed for accurate position estimation at the PSM, while the MOESP method is used to derive the state-space matrices necessary for the KF's operation. By leveraging the JIGSAWS dataset, which includes comprehensive robotic surgical data, we have modeled the PSM's dynamics without relying on prior system knowledge.

Our results indicate that the KF, supported by the MOESP-identified state-space model, achieves high estimation accuracy and low RMSE values under various network conditions, including delays, jitter, and packet loss. Specifically, the KF maintains accuracies above 95\% for all positions (x, y, z), demonstrating its robustness and reliability in the face of network-induced uncertainties. This makes the proposed method highly suitable for critical applications like remote robotic surgery, where precise and real-time position estimation is essential.

The study underscores the effectiveness of combining the KF with the MOESP method to enhance the precision and reliability of remote surgical procedures within the TI framework. Future work will focus on optimizing the KF for more complex network conditions and integrating adaptive filtering techniques and lightweight machine learning algorithms to improve prediction accuracy and adaptability in dynamic surgical environments.

\section*{ACKNOWLEDGMENT}
The Palmer Department Chair and the Richardson Professorship Endowments partially supported the work in this paper.

%\section*{REFERENCES}

\newpage
\newpage
\begin{IEEEbiography}[{\includegraphics[width=1in,height=1.25in,clip,keepaspectratio]{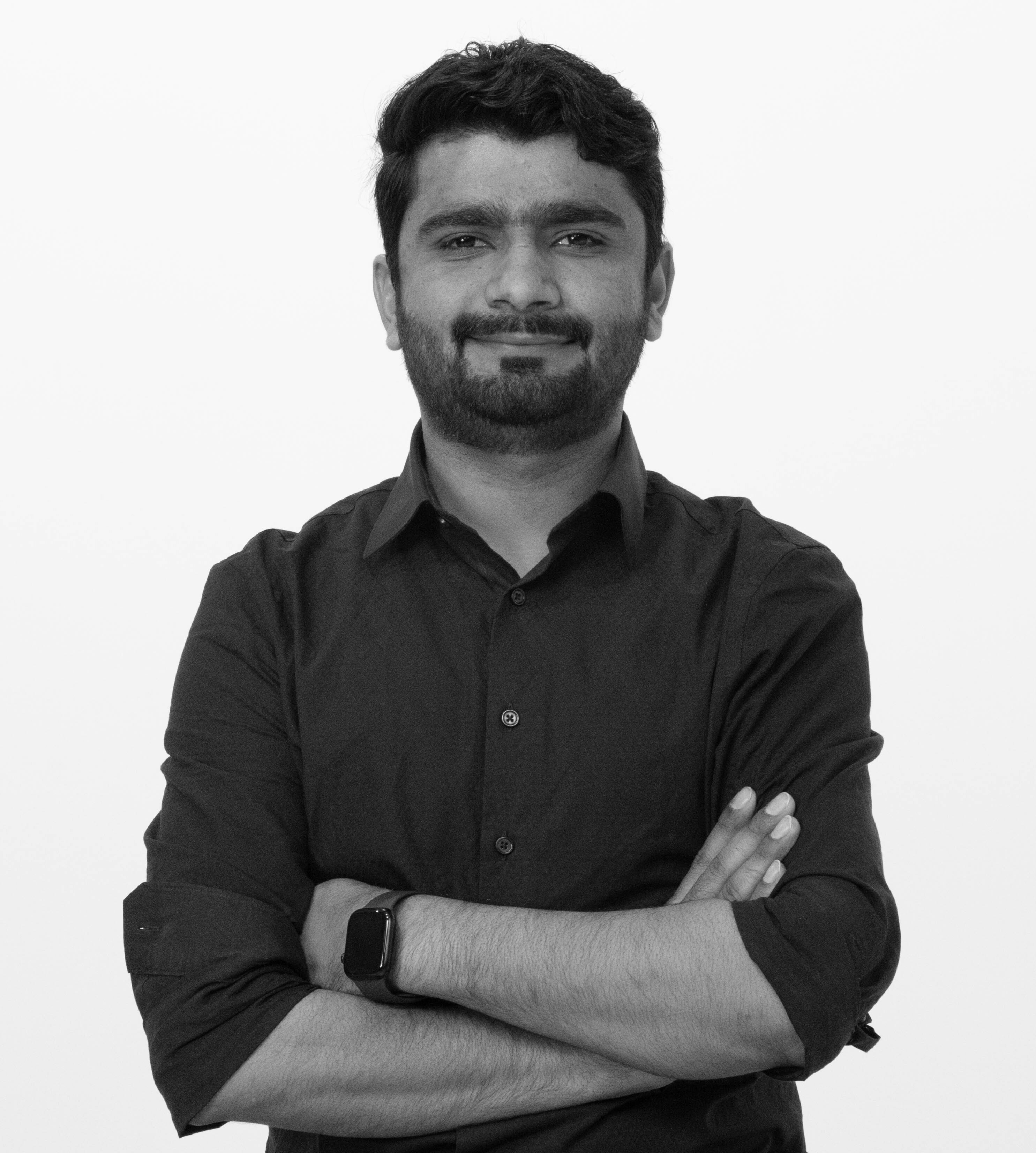}}]
{Muhammad Hanif Lashari }
was born in Sehwan, Sindh, Pakistan. He earned his Bachelor of Engineering (B.E.) in Electronic Engineering in 2016 and his Master of Engineering (M.E.) in Electrical Engineering in 2019, from Mehran University of Engineering \& Technology, Jamshoro, Pakistan. He is pursuing his Ph.D. in Computer Engineering at Iowa State University, Ames, Iowa, USA.
He has held roles as a Lecturer in a public sector university in Pakistan and is currently working as a Graduate Research Assistant in the Department of Electrical and Computer Engineering at Iowa State University. His research interests span the Internet of Things (IoT), Tactile Internet, and Machine Learning, with a specific focus on enhancing prediction in Tactile Internet applications. He has published two journal and conference papers and continues to contribute actively to his field.
Mr. Lashari is also involved in various academic activities beyond his immediate research, aiming to expand his contributions to the field of computer engineering.
\end{IEEEbiography}
\begin{IEEEbiography}[{\includegraphics[width=1in,height=1.25in,clip,keepaspectratio]{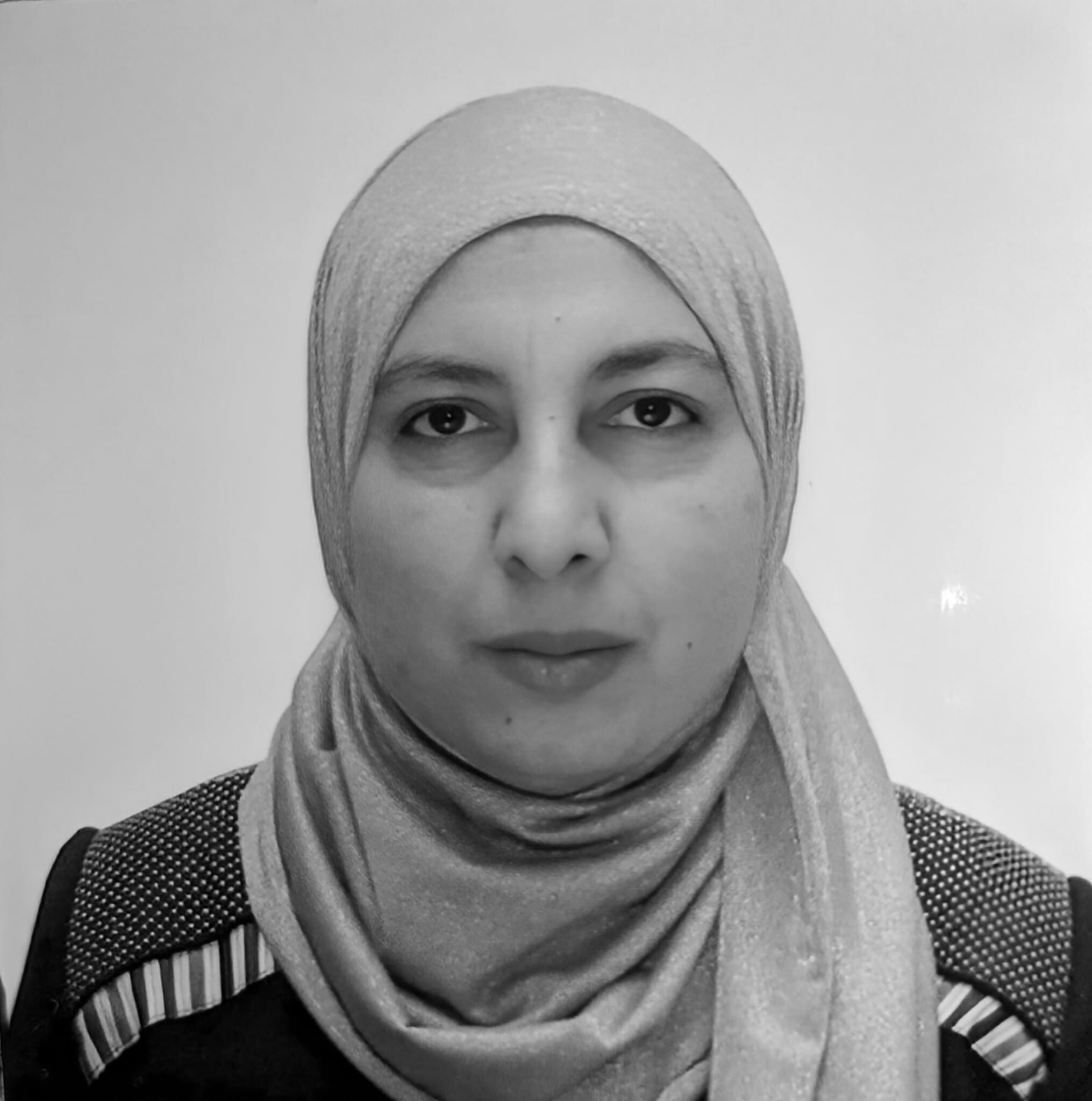}}]
{Wafa Batayneh }was born in Irbid, Jordan. She received her Master's degree and Ph.D. in mechanical engineering from Rensselaer Polytechnic Institute, Troy, NY, USA, in 2005. Her studies focused on Mechatronics systems design.
She has held various positions in academia and research, including summer and fellowship roles, which have contributed to her expertise in robotics and embedded control systems. Currently, she is a Visiting Scholar at the Department of Electrical and Computer Engineering at Iowa State University, Ames, IA, USA, and a Professor at the Mechanical Engineering Department at Jordan University of Science and Technology. She has authored significant publications, including Mechanics of Robotics (Springer, 2010).
Prof. Batayneh is a member of several professional societies. She has received numerous awards for her research and contributions, including national awards.
\end{IEEEbiography}

\begin{IEEEbiography}[{\includegraphics[width=1in,height=1.25in,clip,keepaspectratio]{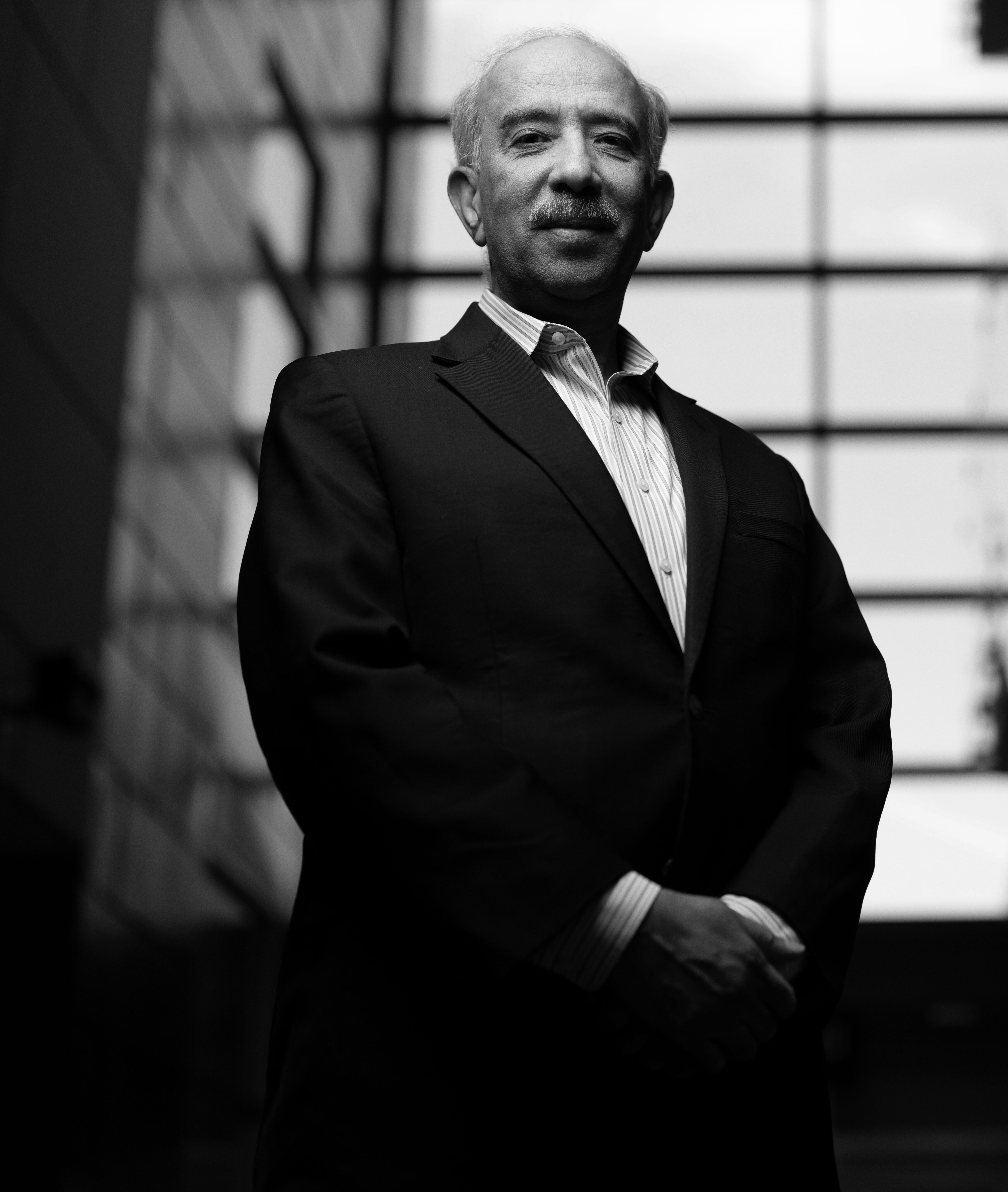}}]
{Ashfaq Khokhar}~(Fellow, IEEE)~currently serves as the Palmer Department Chair of Electrical and Computer Engineering at Iowa State University, a position he has held since January 2017. Previously, he was the Chair of the Department of Electrical and Computer Engineering at the Illinois Institute of Technology from 2013 to 2016. Before that, Khokhar served as a Professor and Director of Graduate Studies in the Department of Electrical and Computer Engineering at the University of Illinois at Chicago.
In 2009, Khokhar was named a Fellow of the Institute of Electrical and Electronics Engineers (IEEE). His accolades also include the NSF Career Award and several best paper awards.
Khokhar's research focuses on context-aware wireless networks, computational biology, machine learning in healthcare, content-based multimedia modeling, retrieval, multimedia communication, and high-performance algorithms. He is renowned for his expertise in developing high-performance solutions for data- and communication-intensive multimedia applications. With over 360 papers published in peer-reviewed journals and conferences, his research has garnered support from the National Science Foundation, the National Institutes of Health, the United States Army, the Department of Homeland Security, and the Air Force Office of Scientific Research.
Khokhar holds a bachelor's degree in electrical engineering from the University of Engineering and Technology in Lahore, Pakistan, a master's degree in computer engineering from Syracuse University, and a Ph.D. in computer engineering from the University of Southern California.
\end{IEEEbiography}

\begin{IEEEbiography}[{\includegraphics[width=1in,height=1.25in,clip,keepaspectratio]{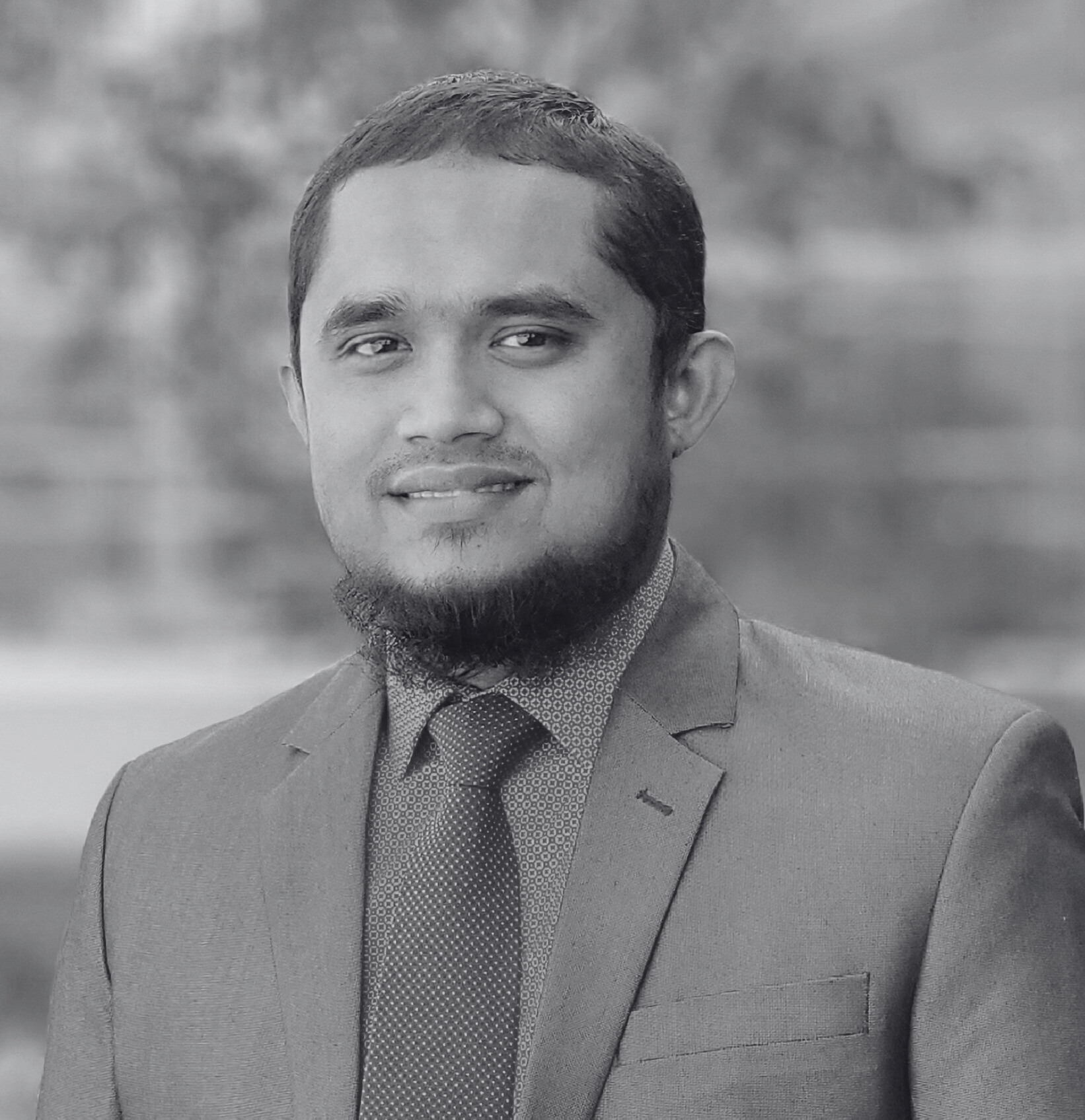}}]
{Shakil Ahmed}~(Member, IEEE)~is an Assistant Teaching Professor in the Department of Electrical and Computer Engineering at Iowa State University. He received his B.S. degree in Electrical and Electronic Engineering from Khulna University of Engineering \& Technology, Bangladesh, in 2014 and his M.S. degree in Electrical Engineering from Utah State University, USA, in 2019. He later earned his Ph.D. in Electrical Engineering and Computer from Iowa State University in a record time of 2 years and 8 months. Ahmed has published numerous research papers in renowned international conferences and journals and received the Best Paper Award at international venues. His research interests include cutting-edge areas such as next-generation wireless communications, GenAI, wireless network design and optimization, unmanned aerial vehicles, physical layer security, Digital Twin for wireless communications, content creation using R.F. signals, and reconfigurable intelligent systems. He is also passionate about engineering education and integrating generative A.I. into learning processes. He has also been a guest editor and reviewer for prestigious journals, including IEEE Transactions on Cognitive Communications and Networking, IEEE Access, IEEE Systems Journal, and IEEE Transactions on Vehicular Technology.
\end{IEEEbiography}

\vfill\pagebreak


\begin{thebibliography}{00}\leftskip1pc
\bibitem{1} Sengupta, Jayasree and Dey, Debasmita and Ferlin, Simone and Ghosh, Nirnay and Bajpai, Vaibhav, ``Accelerating Tactile Internet with QUIC: A Security and Privacy Perspective,'' {\it arXiv preprint arXiv:2401.06657}, 2024.

\bibitem{2} Fitzek, Frank H. P., et al., eds. {\it Tactile Internet: with Human-in-the-Loop}, Academic Press, 2021. [Online]. Available: https://publikationen.bibliothek.kit.edu/1000141406.

\bibitem{3} Kroep, Kees, et al. ``TIM: A Novel Quality of Service Metric for Tactile Internet,'' in {\it Proceedings of the ACM/IEEE 14th International Conference on Cyber-Physical Systems (with CPS-IoT Week 2023)}, pp. 199--208, 2023.
\bibitem{4} Zoltanski, Filip, ``The subjective effect of latency on haptic feedback,'' M.S. thesis, 2023.
\bibitem{5} Beyranvand, Hamzeh, et al. ``Toward 5G: FiWi enhanced LTE-A HetNets with reliable low-latency fiber backhaul sharing and WiFi offloading,'' {\it IEEE/ACM Transactions on Networking}, vol. 25, no. 2, pp. 690--707, 2016.
\bibitem{6} Steinbach, Eckehard, et al. ``Haptic communications,'' {\it Proceedings of the IEEE}, vol. 100, no. 4, pp. 937--956, 2012.
\bibitem{7} Antonakoglou, Konstantinos, et al. ``Toward haptic communications over the 5G tactile Internet,'' {\it IEEE Communications Surveys \& Tutorials}, vol. 20, no. 4, pp. 3034--3059, 2018.
\bibitem{8} Salh, Adeeb, et al. ``A survey on deep learning for ultra-reliable and low-latency communications challenges on 6G wireless systems,'' {\it IEEE Access}, vol. 9, pp. 55098--55131, 2021.

\bibitem{9} Lashari, Muhammad Hanif, Batayneh, Wafa, and Khokhar, Ashfaq, ``Enhancing Precision in Tactile Internet-Enabled Remote Robotic Surgery: Kalman Filter Approach,'' {\it arXiv e-prints}, arXiv:2406.04503, 2024, doi: 10.48550/arXiv.2406.04503.

\bibitem{91} Boabang, Francis, et al. ``A machine learning framework for handling delayed/lost packets in tactile internet remote robotic surgery,'' {\it IEEE Transactions on Network and Service Management}, vol. 18, no. 4, pp. 4829--4845, 2021.

\bibitem{10} Wang, Ziheng and Majewicz Fey, Ann, ``Deep learning with convolutional neural network for objective skill evaluation in robot-assisted surgery,'' {\it International journal of computer assisted radiology and surgery}, vol. 13, pp. 1959--1970, 2018.

\bibitem{29} Jia, Xiaodong, et al. ``Human collective intelligence inspired multi-view representation learning—Enabling view communication by simulating human communication mechanism,'' {\it IEEE Transactions on Pattern Analysis and Machine Intelligence}, 2022.

\bibitem{11} Kazanzides, Peter, et al. ``An open-source research kit for the da Vinci Surgical System,'' in {\it 2014 IEEE international conference on robotics and automation (ICRA)}, pp. 6434--6439, 2014.
\bibitem{12} Fontanelli, Giuseppe Andrea, et al. ``Modelling and identification of the da Vinci research kit robotic arms,'' in {\it 2017 IEEE/RSJ International Conference on Intelligent Robots and Systems (IROS)}, pp. 1464--1469, 2017.
\bibitem{13} Piqué, Francesco, et al. ``Dynamic modeling of the da Vinci research kit arm for the estimation of interaction wrench,'' in {\it 2019 International Symposium on Medical Robotics (ISMR)}, pp. 1--7, 2019.

\bibitem{14} Wang, Yan, et al. ``A convex optimization-based dynamic model identification package for the da Vinci Research Kit,'' {\it IEEE Robotics and Automation Letters}, vol. 4, no. 4, pp. 3657--3664, 2019.

\bibitem{116} Urrea, Claudio and Agramonte, Rayko, ``Kalman filter: historical overview and review of its use in robotics 60 years after its creation,'' {\it Journal of Sensors}, vol. 2021, pp. 1--21, 2021.

\bibitem{216} Ma, Kezhao, et al. ``Review of the Applications of Kalman Filtering in Quantum Systems,'' {\it Symmetry}, vol. 14, no. 12, pp. 2478, 2022.

\bibitem{17} Intuitive Surgical, Inc., ``Da Vinci Surgical System,'' [Online]. Available: https://www.davincisurgery.com/, 2022, Accessed on 29 November 2023.

\bibitem{18} Lefor, Alan Kawarai, et al. ``Motion analysis of the JHU-ISI gesture and skill assessment working set using robotics video and motion assessment software,'' {\it International Journal of Computer Assisted Radiology and Surgery}, vol. 15, pp. 2017--2025, 2020.

\bibitem{19} Ferro, Marco, et al. ``A CoppeliaSim dynamic simulator for the da Vinci Research Kit,'' {\it IEEE Robotics and Automation Letters}, vol. 8, no. 1, pp. 129--136, 2022.
\bibitem{20} Akatsuka, Hiroto and Terada, Masayuki, ``Application of Kalman Filter to Large-scale Geospatial Data: Modeling Population Dynamics,'' {\it ACM Transactions on Spatial Algorithms and Systems}, vol. 9, no. 1, pp. 1--29, 2023.
\bibitem{21} Anh, Nguyen Xuan, et al. ``Towards near real-time assessment of surgical skills: A comparison of feature extraction techniques,'' {\it Computer methods and programs in biomedicine}, vol. 187, pp. 105234, 2020.

\bibitem{22} Varga, Pal, et al. ``5G support for industrial IoT applications—challenges, solutions, and research gaps,'' {\it Sensors}, vol. 20, no. 3, pp. 828, 2020.

\bibitem{23} Malekzadeh, Mina, ``Performance prediction and enhancement of 5G networks based on linear regression machine learning,'' {\it EURASIP Journal on Wireless Communications and Networking}, vol. 2023, no. 1, pp. 74, 2023.
\bibitem{24} Fettweis, Gerhard P., ``The tactile internet: Applications and challenges,'' {\it IEEE vehicular technology magazine}, vol. 9, no. 1, pp. 64--70, 2014.


\bibitem{b25} Jamaludin, I. W., Wahab, N. A., Khalid, N. S., Sahlan, S., Ibrahim, Z., \& Rahmat, M. F. (2013, March). N4SID and MOESP subspace identification methods. In 2013 IEEE 9th international colloquium on signal processing and its applications (pp. 140-145). IEEE.

\bibitem{b26} Robles, A. (2017). MOESP-and-N4SID [GitHub repository]. GitHub. https://github.com/Alro10/MOESP-and-N4SID


\bibitem{28} Pokhrel, Shiva Raj, et al. ``Towards enabling critical mMTC: A review of URLLC within mMTC,'' {\it IEEE access}, vol. 8, pp. 131796--131813, 2020.

\bibitem{29} Shi, Chang, Yi Zheng, and Ann Majewicz Fey. "Recognition and prediction of surgical gestures and trajectories using transformer models in robot-assisted surgery." 2022 IEEE/RSJ International Conference on Intelligent Robots and Systems (IROS). IEEE, 2022.

\bibitem{31} Weerasinghe, Keshara, et al. "Multimodal Transformers for Real-Time Surgical Activity Prediction." arXiv preprint arXiv:2403.06705 (2024).

\bibitem{30} Xiong, Liang, Xi Chen, and Jeff Schneider. "Direct robust matrix factorizatoin for anomaly detection." 2011 IEEE 11th International Conference on Data Mining. IEEE, 2011.


%\bibitem{25} Rao, Y., et al. ``New services \& applications with 5G ultra-reliable low latency communication,'' {\it 5G Americas, ellevue, WA, USA, Tech. Rep}, 2018.
%\bibitem{26} 3GPP, ``Study on Physical Layer Enhancements for NR Ultra-Reliable and Low Latency Case (URLLC),(Release 16),'' {\it 3GPP TR 38.824 V16. 0.0}, 2019.
%\bibitem{27} Li, Yuke, et al. ``Smart choice for the smart grid: Narrowband Internet of Things (NB-IoT),'' {\it IEEE Internet of Things Journal}, vol. 5, no. 3, pp. 1505--1515, 2017.



\end{thebibliography}
\end{document}